%% file: main.tex
\pdfoutput=1

\documentclass[11pt]{article}
\usepackage[]{packages/emnlp2021}

\usepackage{times}
\usepackage{latexsym}

\usepackage[T1]{fontenc}

\usepackage[utf8]{inputenc}

\usepackage{microtype}

\usepackage{epsfig}
\usepackage{graphicx}
\usepackage{amsmath}
\usepackage{amssymb}
\usepackage{hyperref}
\usepackage{lipsum}
\usepackage{xcolor}
\usepackage[margin=0.6cm,justification=centering]{caption}
\usepackage{listings}
\title{Optimal Size-Performance Tradeoffs: Weighing PoS Tagger Models}

\author{Magnus Jacobsen* \\
  IT University of Copenhagen \\
  \texttt{mjac@itu.dk} \\\And
  Mikkel H. Sørensen* \\
  IT University of Copenhagen \\
  \texttt{mhso@itu.dk} \\\And
  Leon Derczynski \\
  IT University of Copenhagen \\
  \texttt{leod@itu.dk}}

\date{}

\begin{document}

\maketitle

\let\thefootnote\relax\footnotetext{* Joint first authors}

\input{sections/0_abstract}
\input{sections/1_introduction}
\input{sections/2_related_work}
\input{sections/3_method}
\input{sections/4_results}
\input{sections/5_discussion}
\input{sections/6_conclusion}
\input{sections/8_references}

\end{document}

%% file: sections/0_abstract.tex
\begin{abstract}
Improvement in machine learning-based NLP performance are often presented with bigger models and more complex code. This presents a trade-off: better scores come at the cost of larger tools; bigger models tend to require more during training and inference time. We present multiple methods for measuring the size of a model, and for comparing this with the model's performance.

In a case study over part-of-speech tagging, we then apply these techniques to taggers for eight languages and present a novel analysis identifying which taggers are size-performance optimal. Results indicate that some classical taggers place on the size-performance skyline across languages. Further, although the deep models have highest performance for multiple scores, it is often not the most complex of these that reach peak performance.
\end{abstract}

%% file: sections/1_introduction.tex
\section{Introduction}
State of the art machine learning has moved from shallow to deep and more complex models. The upwards trend of model complexity has brought with it an increase in hardware demands, including more memory, disk space, and computational power. The full instance of OpenAI’s deep learning language model \textit{GPT-3}, trained on a compressed dataset of 570GB, boasts 175 billion model parameters~\cite{Brown2020LanguageLearners}; Switch-C, a trillion~\cite{fedus2021switch}. 

However, not all applications of machine learning are on hardware capable of running these models. Devices such as smartphones, hearing aids, or smart speakers are too constrained. Natural Language Processing (NLP) models are important for the devices mentioned above, from voice assistants on smart phones and speakers, to going beyond sound amplification in hearing aids. Using NLP in this context requires considering model size, and balancing the tradeoff between size and task accuracy. And indeed, larger language models can bring their own drawbacks~\cite{bender2021dangers}.

Determining model size requires metrics. Measuring model filesize in bytes is flawed: some models are compressed; some are uncompressed; some are compressed badly; others embed important functionality in their code instead~\cite{barf}; others still impose much larger space constraints at inference time than their dormant, on-disk size indicates. Different starting points also offer varying model size -- a custom image might only need a script of a few lines to perform a complex task, that would have many more dependencies on a standard operating system installation. To compare these, common metrics and definitions are required that address these problems.

Our goal is to critically examine the relationship between size complexity and prediction performance of NLP models. We do not examine the time required for training or for inference, only the size of models. We take part-of-speech (PoS) tagging as a case study. PoS tagging is a well-defined problem, with well-defined sets of inputs and labeled outputs. 
Readily measurable output benefits performance measurements and comparisons. 
PoS tagging is also no different from the trend in machine learning, where \textit{better} has become coupled to \textit{bigger}.

Our research questions are: \textbf{RQ1} How can we measure the size of an NLP tool in a reproducible, comparable way? \textbf{RQ2} How can we determine which tools are size-performance optimal?

In relation to prediction performance, we probe how well taggers handle a diverse range of languages. This will test their robustness and ability to generalise when presented with different inputs. In relation to size we intend to investigate models of varying complexity and parameter count. This will show how PoS taggers of different size perform. We explore various PoS tagging approaches and methods for minimizing the size of machine learning models. Further, we evaluate the prediction accuracy of various PoS taggers in comparison to their size, and finally discuss these findings in relation to further research to model size optimality.




%% file: sections/2_related_work.tex
\section{Related Work}\label{sec:relatedwork}
As machine learning models have grown, more research has investigated optimizing model size. 

Compression has shown that large networks rarely make full use of the space they take up~\cite{Han2016DeepCoding}. One form of model compression is to use limit precision for parameters. \newcite{Gupta2015DeepPrecision} experimented with limited precision with fixed-point numbers. Similarly, \newcite{Wang2018TrainingNumbers} experimented with quantized limited precision floating point numbers. Both compression schemes needed stochastic rounding to preserve accuracy. 
The extreme cases of limited precision are binary or ternary networks, where weights are either -1 0 or 1. This was explored for LSTM networks~\cite{Ardakani2019LearningWeights} and CNNs~\cite{Ignatov2020ControllingNetwork}. 



The advent of pre-trained embeddings and transformer models, such as BERT~\cite{Devlin2019BERT:Understanding}, prompted research into minimizing their size. Word embeddings are an integral part in deep learning PoS taggers, but large. Several works have explored embedding compression~\cite{Chen2016CompressingRepresentations, Tissier2019Near-losslessEmbeddings,Kim2020AdaptiveEmbeddings, Melas-Kyriazi2020Generation-distillationSettings,Liao2020EmbeddingQuantization}.


Another form of resource-constrained NLP is with low-resource languages or datasets. Research into unsupervised PoS tagging on low resource languages has been done by \newcite{Buys2016Cross-lingualLanguages,Cardenas2019ALanguages}. \newcite{Ezen-Can2020ACorpus} evaluated the performance of BERT on a small dataset.

The most similar research to ours is \newcite{Szymanski2020IsProcessing}, who use a Bayesian approach to evaluate six PoS taggers across two datasets. However, they did not conduct size measurements or analysis, and only covered English.

%% file: sections/3_method.tex
\section{Method}
We evaluated ten PoS taggers across eight languages, to see the size vs. accuracy trade-off of classical and contemporary taggers, measuring performance using two accuracy metrics and three size metrics.

\subsection{Taggers}
We reviewed and experimented on five taggers that utilize classical non-neural machine learning techniques. We did the same for five modern taggers utilizing deep learning. The majority of these taggers achieved state-of-the-art performance at the time of their inception.

We modified the code for some taggers, to make them compatible with our testing platform. We also changed how some of them parsed input to make them compatible with testing data. For some taggers, we created an inference function, as it was missing. We did not alter the model architecture in any way for any of the taggers.

\subsubsection{Classical Taggers}
\paragraph{Brill Tagger}
Rule-based tagger~\cite{Brill1992ATagger}. Achieved a state-of-the-art accuracy for its time of 95\% on the English Brown Corpus.

\paragraph{Trigrams'n'Tags (TnT)}
Hidden Markov Model (HMM)~\cite{Brants2000TnTTagger}. Achieved a state-of-the-art accuracy for its time of 96.7\% on the English WSJ PTB~\cite{Marcus1993BuildingTreebank}.

\paragraph{Stanford Tagger}
Uses maximum-entropy approximation and a cyclic dependency network~\cite{Toutanova2003Feature-richNetwork}. Scored state-of-the-art accuracy at the time: 97.24\% on PTB.

\paragraph{SVMTool}
Utilizes Support Vector Machines and achieved an accuracy of 97.16\% on the Penn Treebank corpus~\cite{Gimenez2004SVMTool:Machines}. We use the version written in Perl.

\paragraph{HMM (NLTK)}
First-order HMM from the Natural Language Toolkit (NLTK)~\cite{Bird2002NLTKToolkit}.

\subsubsection{Deep Learning Taggers}
All of the modern taggers we investigated utilized recurrent neural network (RNN) architectures. Specifically, LSTM networks are utilized by all of them to boost state-of-the-art accuracy. All the taggers also employ either word or subword embeddings, or a combination of these. Embeddings and LSTM networks contain many parameters and therefore require more memory and disk space to use than classical approaches.

\paragraph{BiLSTM (Plank)}
Uses an auxiliary loss function to handle rare words~\cite{Plank2016MultilingualLoss}. Reached state-of-the-art scores for its time on 22 languages.

\paragraph{BiLSTM (Yasunaga)}
Employs adversarial training for regularization and a conditional random field (CRF) for prediction~\cite{Yasunaga2018RobustTraining}. Achieved competitive results across different languages, including an accuracy of 97.58\% on PTB.

\paragraph{Flair}
Uses a novel embedding form, ``contextual string embeddings"~\cite{Akbik2018ContextualLabeling}. Achieved a state-of-the-art accuracy for its time of 97.82\%.

\paragraph{Meta-BiLSTM}
Combines three separately trained BiLSTM neural networks~\cite{Bohnet2018MorphosyntacticEncodings}. Achieved state-of-the-art accuracy for its time of 97.96\% on the PTB corpus.

\paragraph{BiLSTM (Heinzerling)}
Uses a combination of sub-word embeddings, including the BERT transformer model~\cite{Heinzerling2020SequenceEvaluation}. Gave competitive accuracy across 27 languages.

\subsection{Metrics} \label{meth:metrics}
Measuring the size and accuracy trade-off of machine learning models is not a trivial issue. For different applications, different size constraints might apply. To get a broad view of performance, we measured each tagger with respect to three size metrics and two accuracy metrics. We measured token accuracy and sentence accuracy. Below, we outline the three size metrics, and how we measure them.

\paragraph{Memory}
We measured memory usage during inference on test data. For each tagger, we ran inference in an isolated process and polled the memory usage twice every second. The reported memory usage for a tagger is the average usage across these measurements. This should provide a realistic view on how much memory a tagger would consume in a typical application.

\paragraph{Model Size (Uncompressed)}
We measured the disk space required by a tagger by how many bytes its trained parameters took up on disk. We also included any pre-trained embeddings for the tagger in this measurement. These were included because embeddings can be seen as additional trained parameters for the model.

\paragraph{Model Size (Compressed)}
We compressed necessary model files for a tagger to the lossless .xz archiving format. This is not to be confused with model compression as mentioned in section~\ref{sec:relatedwork}. This gives a naive lower bound for the potential size of a trained model, without altering its architecture.

\subsection{Datasets}
To test the multilingual capabilities of the taggers, we chose a diverse selection of languages. These languages span different families and contain different morphological characteristics. The languages used are in table~\ref{tab:languages}.

\begin{table}[h]
\centering
\begin{footnotesize}
\setlength\tabcolsep{5.5pt}
\begin{tabular}{l|cccc}
\hline
 & Code & Family & Branch \\
 \hline
Arabic & ar & Afro-Asiatic & Semitic \\
Chinese & zh & Sino-Tibetan & Sinitic \\
Danish & da & Indo-European & Germanic \\
English & en & Indo-European & Germanic \\
Hindi & hi & Indo-European & Indo-Iranian \\
Russian & ru & Indo-European & Balto-Slavic \\
Spanish & es & Indo-European & Italic \\
Turkish & tr & Turkic & Common Turkic \\
\hline
\end{tabular}
\end{footnotesize}
\caption{Grouping of languages. Codes are ISO 639-1.}
\label{tab:languages}
\end{table}

We used Universal Dependencies (UD) 2.6 datasets~\cite{Nivre2020UniversalCollection}. UD provides uniformly labelled datasets for a large array of languages. The languages together with the chosen dataset for each, are in table \ref{tab:datasets}. The Chinese dataset uses Traditional Chinese characters. The split were pre-determined by the UD datasets.

\begin{table}[h]
\centering
\setlength\tabcolsep{5.5pt}
\begin{footnotesize}
\begin{tabular}{l|cccc}
\hline
\; & Dataset & Tokens & Sentences & Splits\\
\hline
Arabic      & PADT  & 242K & 7,664 & 80/10/10\\
Chinese     & GSD   & 123K & 4,997 & 80/10/10\\
Danish      & DDT   & 101K & 5,512 & 80/10/10\\
English     & GUM   & 113K & 5,961 & 72/14/14\\
Hindi       & HDTB  & 352K & 16,647 & 80/10/10\\
Russian     & GSD   & 98K  & 5,030 & 76/12/12\\
Spanish     & AnCora & 548K & 17,680 & 80/10/10\\
Turkish     & IMST  & 56.4K  & 5,635 & 66/17/17\\
\hline
\end{tabular}
\end{footnotesize}
\caption{Datasets used for experiments. Splits are approximate percentages of total tokens for training, development, and testing sets, respectively.}
\label{tab:datasets}
\end{table}

\paragraph{Data Curation}
Some taggers expected input data structured differently from the files found in UD. To deal with this issue, we created versions of the datasets where sentences were merely token and tag pairs. We also removed non-tagged multitoken words from the original \textit{.conllu} files.  None of the taggers expected multitoken words. These were introduced with version 2 of UD.

For SVMTool we made some minor data modifications. We had to guarantee that every sentence ended with a full stop.

\paragraph{Pre-Trained Embeddings}
Some taggers used embeddings, often trained unsupervised over a large corpus. They encapsulate contextual word information, represented as fixed-size vectors.

Polyglot word embeddings~\cite{Al-Rfou2013Polyglot:NLP} with a dimensionality of 64 were used for BiLSTM (Plank), BiLSTM (Yasunaga), and Meta-BiLSTM. BPEmb byte-pair embeddings~\cite{Heinzerling2019BPEMB:Languages} with 100 dimensions were used for the BERT-BPEmb tagger. A combination of Polyglot and FastText embeddings~\cite{Joulin2017BagClassification} was used with Flair, the latter having 300 dimensions.

\subsection{Experiment Setup}
To conduct experiments with taggers written in different programming languages, we needed a flexible testing platform. We built this platform using Python.

The specifications for the hardware used for experiments are: dual Intel Xeon Platinum 8176z CPU; 503GB memory; Ubuntu 18.04 (64-bit); Python 3.8.5.

%% file: sections/4_results.tex
\section{Results}
Here we report measurement results from our experiments, for both accuracy and size metrics. We then link these metrics to investigate tradeoffs.

\subsection{Accuracy Measurements}
The accuracy measurements from our experiments are found in table \ref{tab:accuracies} and \ref{tab:sent-accuracies}. The first table shows results when predicting individual tags for tokens. The second table shows accuracy for predicting complete sentences. All measurements were done from taggers predicting on unseen test datasets.

\begin{table*}[!htb]
\footnotesize
\setlength\tabcolsep{5.5pt}
\begin{tabular}{l|cccccccccc|c}
\hline
 & Brill & TnT & SVMTool & \begin{tabular}[c]{@{}c@{}}Stanford\\ Tagger\end{tabular} & HMM & \begin{tabular}[c]{@{}c@{}}BiLSTM/\\ Plank\end{tabular} & \begin{tabular}[c]{@{}c@{}}BiLSTM/\\ Yasunaga\end{tabular} & Flair & \begin{tabular}[c]{@{}c@{}}Meta-\\ BiLSTM\end{tabular} & \begin{tabular}[c]{@{}c@{}}BERT-\\ BPEmb\end{tabular} & Avg. \\
\hline
ar & 92.36 & 90.49 & 95.66 & 95.73 & 92.25 & 96.43 & 96.55 & \textbf{96.68} & 95.80 & 96.27 & 94.82 \\
zh & 83.50 & 80.59 & 91.28 & 91.84 & 83.04 & 93.51 & 93.17 & 91.47 & 92.95 & \textbf{95.08} & 89.64 \\
en & 84.07 & 80.15 & 93.54 & 93.64 & 83.73 & 95.50 & 95.60 & \textbf{96.29} & 95.34 & 94.52 & 91.24 \\
es & 94.50 & 92.18 & 97.93 & 98.12 & 94.11 & 98.50 & 98.19 & \textbf{98.81} & 98.22 & 98.64 & 96.92 \\
da & 86.08 & 80.76 & 94.60 & - & 86.15 & 96.93 & 96.08 & \textbf{97.36} & 95.96 & 97.17 & 83.11 \\
hi & 92.75 & 91.10 & 96.16 & - & 92.09 & 97.10 & 96.50 & \textbf{97.32} & 96.23 & 96.82 & 85.61 \\
ru & 80.00 & 71.26 & 95.10 & - & 79.74 & 97.36 & 96.49 & \textbf{97.55} & 97.00 & 97.30 & 81.18 \\
tr & 78.33 & 70.95 & 92.33 & - & 78.08 & 94.90 & 91.86 & \textbf{96.02} & 93.83 & 95.34 & 79.16 \\
\hline
Avg. & 86.45 & 82.19 & 94.58 & - & 86.15 & 96.28 & 95.55 & \textbf{96.44} & 95.67 & 96.39 & - \\
\hline
\end{tabular}
\caption{Token accuracy on test set, for all taggers on all supported languages. Numbers are in percent. \textit{\textbf{Bold is best row score.}}}
\label{tab:accuracies}
\end{table*}

\begin{table*}[!htb]
\setlength\tabcolsep{5.5pt}
\begin{footnotesize}
\begin{tabular}{l|cccccccccc|c}
\hline
 & Brill & TnT & SVMTool & \begin{tabular}[c]{@{}c@{}}Stanford\\ Tagger\end{tabular} & HMM & \begin{tabular}[c]{@{}c@{}}BiLSTM/\\ Plank\end{tabular} & \begin{tabular}[c]{@{}c@{}}BiLSTM/\\ Yasunaga\end{tabular} & Flair & \begin{tabular}[c]{@{}c@{}}Meta-\\ BiLSTM\end{tabular} & \begin{tabular}[c]{@{}c@{}}BERT-\\ BPEmb\end{tabular} & Avg. \\
\hline
r & 23.68 & 17.50 & 37.79 & 38.53 & 23.24 & 44.26 & 46.00 & \textbf{46.76} & 37.50 & 44.56 & 35.98 \\
zh & 6.40 & 2.40 & 22.60 & 26.80 & 5.60 & 32.60 & 29.60 & 24.40 & 26.00 & \textbf{39.60} & 21.60 \\
en & 17.30 & 12.13 & 43.03 & 45.17 & 16.63 & 52.36 & 54.00 & \textbf{59.89} & 53.15 & 48.31 & 40.20 \\
es & 28.76 & 18.07 & 60.37 & 64.73 & 25.62 & 68.74 & 64.41 & \textbf{74.32} & 64.44 & 72.57 & 54.20 \\
da & 17.52 & 8.50 & 46.02 & - & 17.52 & 61.06 & 57.40 & \textbf{68.32} & 55.22 & 65.84 & 39.74 \\
hi & 33.43 & 24.23 & 56.35 & - & 28.98 & 63.12 & 59.13 & \textbf{67.16} & 54.22 & 62.65 & 44.93 \\
ru & 10.32 & 2.33 & 48.42 & - & 9.32 & 69.05 & 59.33 & \textbf{69.22} & 63.56 & 64.89 & 39.64 \\
tr & 19.02 & 10.17 & 51.88 & - & 18.62 & 62.46 & 49.33 & \textbf{69.89} & 57.68 & 65.11 & 40.42 \\
\hline
Avg. & 19.55 & 11.92 & 45.81 & - & 18.19 & 56.71 & 52.40 & \textbf{60.00} & 51.47 & 57.94 & - \\
\hline
\end{tabular}
\end{footnotesize}
\caption{Sentence accuracy on test set, for all taggers on all supported languages. Numbers are in percent.}
\label{tab:sent-accuracies}
\end{table*}

\paragraph{Token Accuracy}
Results for token accuracy are in Table~\ref{tab:sent-accuracies}. Deep learning models performed better almost across the board. The network that performed best on average, across all languages, was Flair. These results are not surprising. Flair uses pre-trained embeddings with more dimensions than the other deep learning taggers.

Flair performed best on seven individual languages. The only language it performed poorly on was Chinese. It had the lowest accuracy among the deep learning models. Chinese is the language with the 3rd lowest average accuracy among all taggers. Even still, Flair's low performance is remarkable. This could be because its architecture is built around splitting words and tokens into sequences of characters~\cite{Akbik2018ContextualLabeling}. Chinese is a logographic languages, which makes this difficult. BERT-BPEmb performs better than all other taggers on Chinese. This suggests that the attention architecture found in its
BERT component~\cite{Devlin2019BERT:Understanding} is better than BiLSTMs at learning Chinese language form.

\begin{figure}[!h]
    \centering
    \includegraphics[width=0.95\columnwidth]{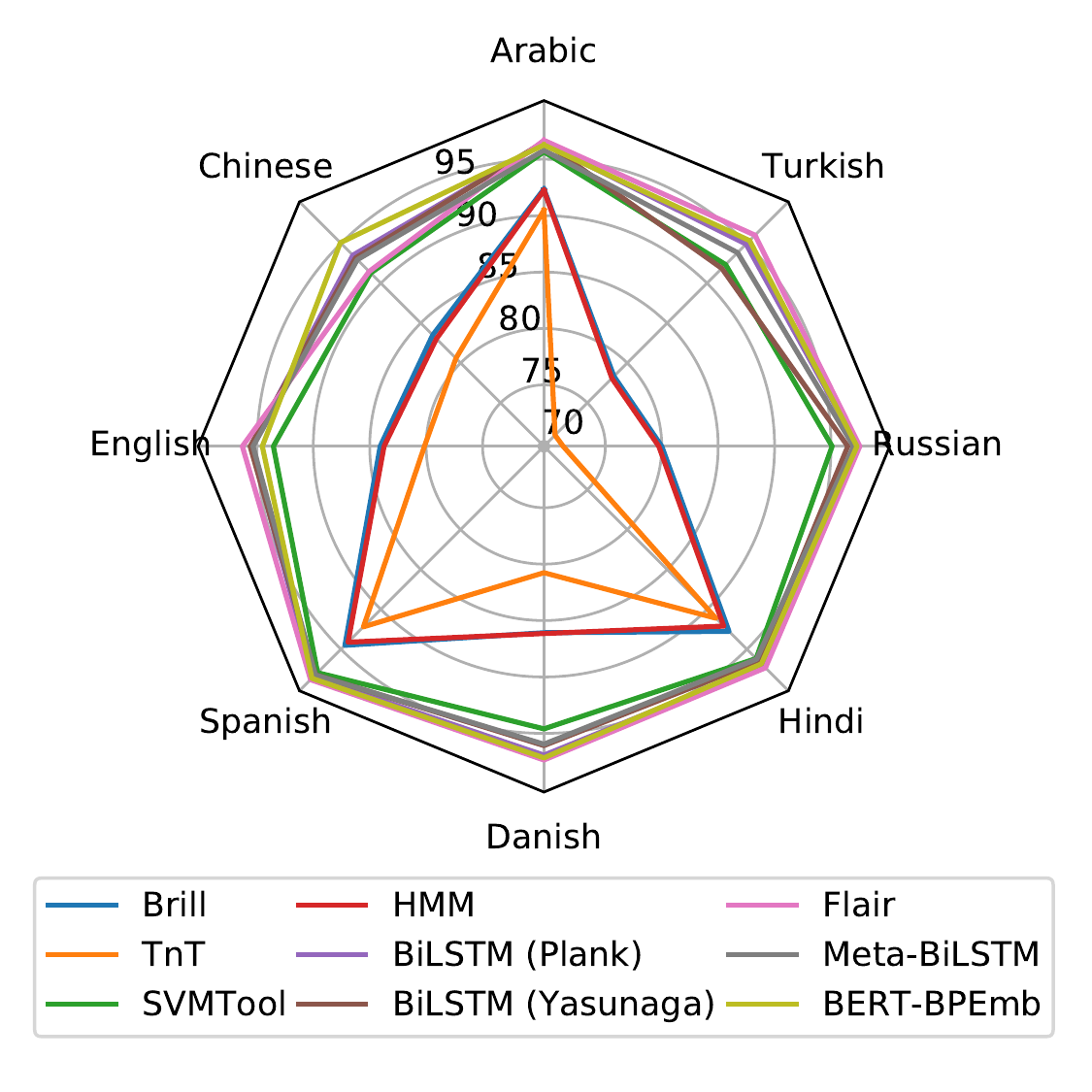}
    \caption{Taggers token performance for each language. Stanford Tagger excluded.}
    \label{fig:spiderweb}
\end{figure}

Among classical taggers, both SVMTool and the Stanford Tagger perform well. They are behind all the deep learning models in performance, but lie closer to these than to the other classical taggers. Figure~\ref{fig:spiderweb} shows how competitive the accuracy of SVMTool is in comparison to the deep learning taggers. The gap in performance for Brill, HMM, and TnT is considerably larger.

\paragraph{Sentence Accuracy}
Results for sentence accuracy are in table \ref{tab:sent-accuracies}. Here, the difference between classical and deep learning taggers is even more pronounced. The performance of the Stanford Tagger and SVMTool are far better than the other classical taggers. Again, the deep learning taggers perform better across the board. Flair has the best performance on the same seven languages as with token accuracy. It also has the highest average accuracy across all languages. As with token accuracy, BERT-BPEmb has best performance on Chinese.

Sentence accuracy provides another perspective on how versatile and robust a tagger is. If a given tagger performs well in token accuracy, but poorly in sentence accuracy, it might not grasp the context of words and tags in sentences. In that case, it has merely learned what token generally corresponds to which tag.

\subsection{Size Measurements}
Size measurements across the three chosen metrics are in Table~\ref{tab:sizes}. These are averaged across results for all languages.

\begin{table}[h]
\centering
\setlength\tabcolsep{5.5pt}
\begin{footnotesize}
\begin{tabular}{l|cccc}
\hline
 & Memory & Model & \begin{tabular}[c]{@{}c@{}}Model\\ Compr.\end{tabular} \\
\hline
Brill               & 1.04e5    & 1.99e3    & 1.78e2        \\
TnT                 & 1.99e5    & 6.85e2    & 1.32e2        \\
SVMTool             & 5.77e4    & 4.03e3    & 7.37e2        \\
Stanford Tagger     & 2.92e5    & 5.42e3    & 2.01e3        \\
HMM                 & 1.03e5    & 4.45e2    & 1.30e2        \\
BiLSTM/ Plank       & 1.18e6    & 1.46e5    & 4.87e4        \\
BiLSTM/ Yasunaga    & 5.08e6    & 4.10e4    & 3.37e4        \\
Flair               & 5.50e6    & 7.96e5    & 4.68e5        \\
Meta-BiLSTM         & 2.74e7    & 2.88e5    & 2.60e5        \\
BERT-BPEmb          & 6.80e7    & 1.46e6    & 1.33e6        \\
\hline
\end{tabular}
\end{footnotesize}
\caption{Average size after training, across all languages for each tagger. All sizes are in kilobytes.}
\label{tab:sizes}
\end{table}

\paragraph{Memory}
SVMTool uses the least memory during inference of all taggers. This is due to its reliance on lightweight SVM techniques. BiLSTM (Plank) is the least memory intensive of the deep learning taggers. This is likely due to its machine learning framework DyNet. This framework is more lightweight than TensorFlow or PyTorch which is used by the three largest models. BiLSTM (Yasunaga) use Theano, which is also relatively lightweight. We did not limit available memory for any of the models. Putting constraints on the available memory for the taggers might result in lower numbers. I.e. these figures should be seen as an upper bound.

\paragraph{Model Size (Uncompressed)}
Using this metric, the amount of parameters for the classical taggers are magnitudes smaller than for the deep learning models. The smallest amount belongs to the Stanford Tagger. This is an outlier because it contains hard-coded language-specific features in its code base. This enables the small size on disk for the stored parameters. BiLSTM (Yasunaga) is the most lightweight of the deep learning models. Plank is larger due to storing its parameters in a less efficient format. It is worth remembering that embeddings are included in model size. This factors into the larger sizes of the deep learning models.

\paragraph{Model Size (Compressed)}
SVMTool and BiLSTM (Plank) gains the most from lossless xz compression. This is because their trained parameters are stored in plain text. The size rankings compared to the previous metric is otherwise unchanged.

\subsection{Size-Accuracy Efficiency} \label{res:efficiency}
\begin{figure*}
    \centering
    \includegraphics[width=1\textwidth]{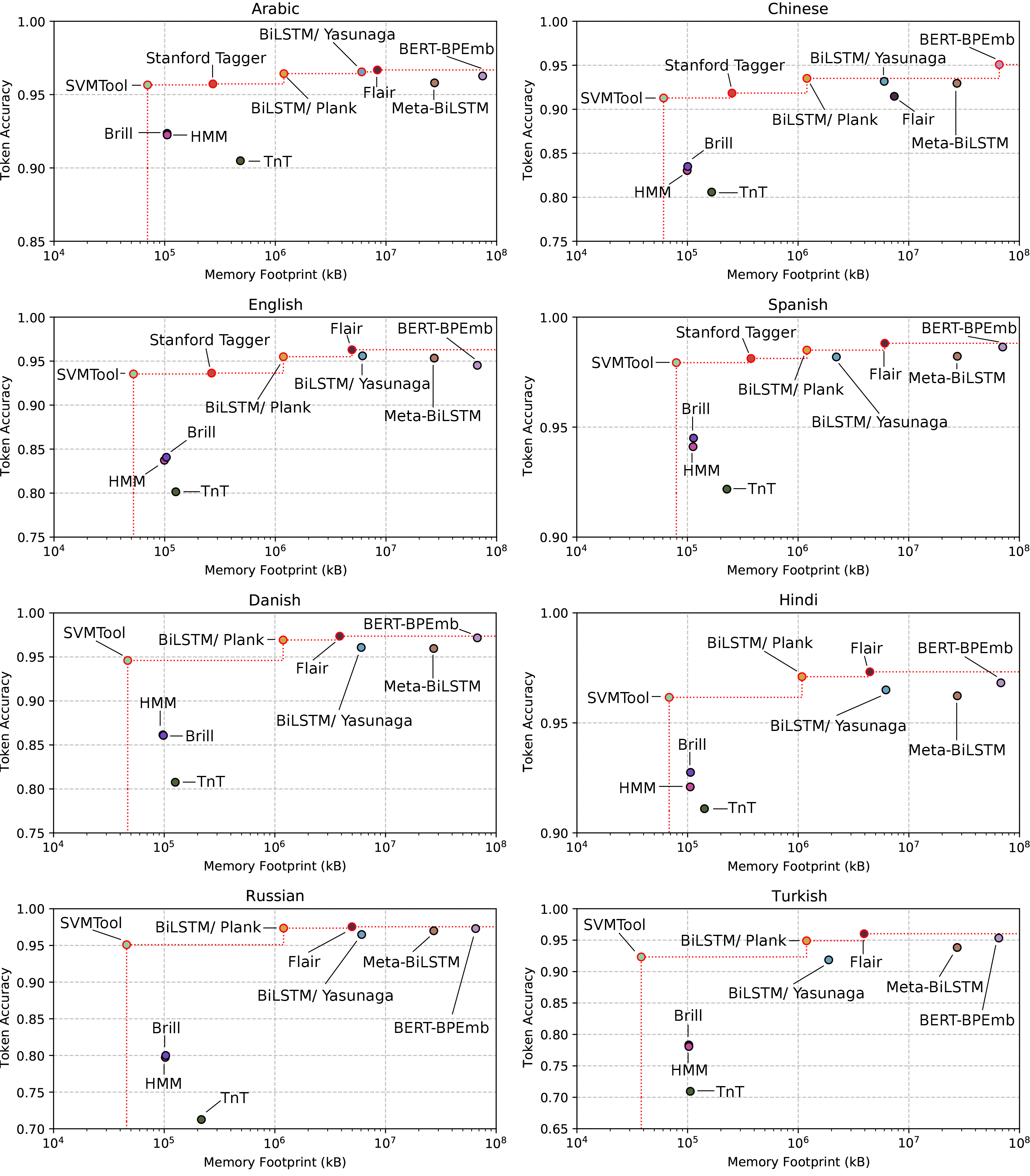}
    \caption{Skyline plots of taggers for all eight languages showing token accuracy vs. memory usage}
    \label{fig:skyline_memory_vs_acc}
\end{figure*}

The eight plots in figure \ref{fig:skyline_memory_vs_acc} show memory usage during inference on the x-axis, and token accuracy on the y-axis. The figure includes plots with measurements across all eight languages for all taggers. The Stanford Tagger is included for the languages it supports. The red dotted line in each plot represents the \textit{skyline}~\cite{Borzsonyi2001TheOperator}. This indicates the measured optimum between memory usage and accuracy. The skyline intersects the optimal taggers. The taggers below the skyline are dominated by better optimums. 

\begin{figure*}[htb]
    \centering
    \includegraphics[width=0.9\textwidth]{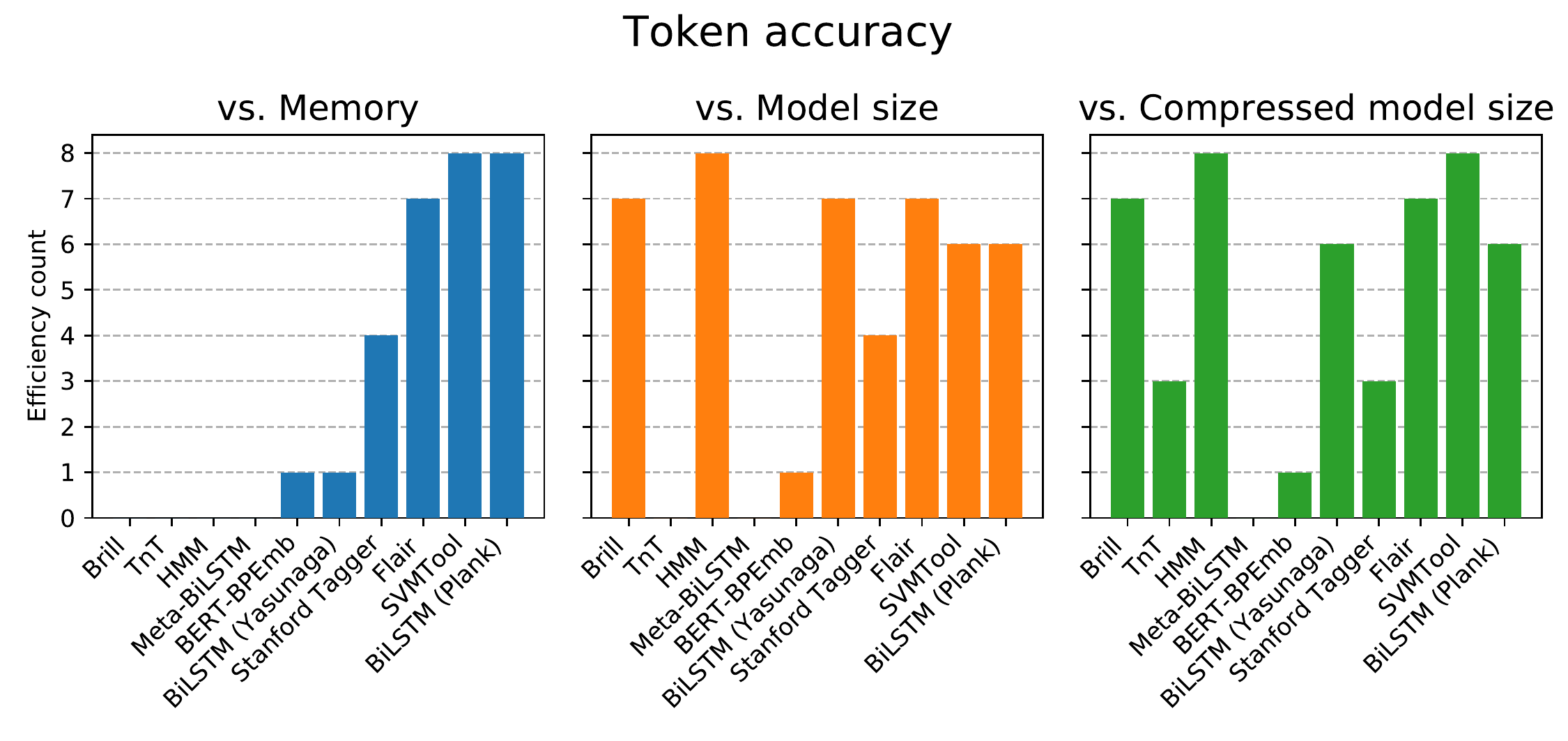}
    \caption{Token accuracy optimality: number of languages for which each tagger was on skyline}
    \label{fig:skyline_rankings_token}
\end{figure*}

\begin{figure*}[htb]
    \centering
    \includegraphics[width=0.9\textwidth]{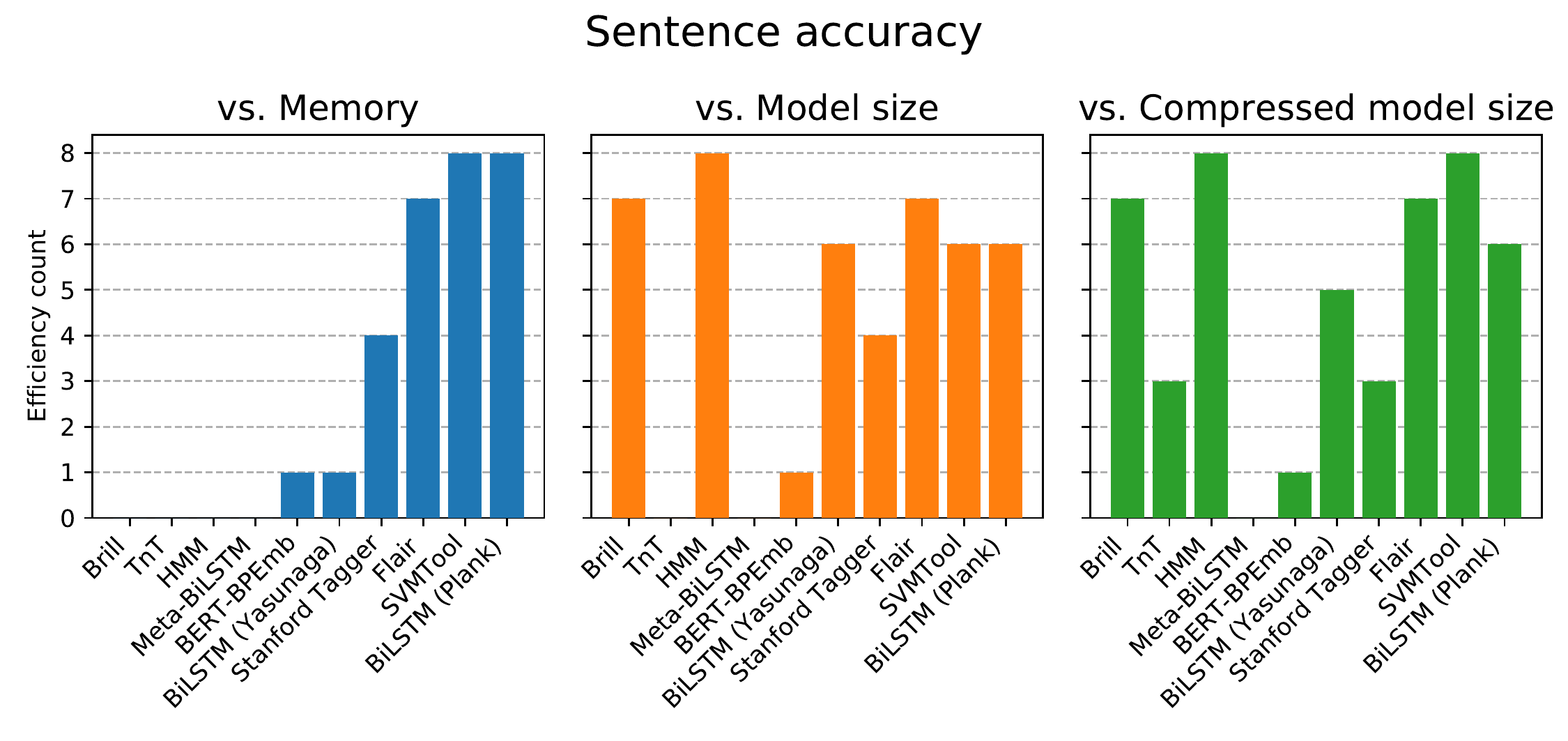}
    \caption{Sentence accuracy optimality: number of languages for which each tagger was on skyline}
    \label{fig:skyline_rankings_sentence}
\end{figure*}

Figure \ref{fig:skyline_rankings_token} and \ref{fig:skyline_rankings_sentence} summarize how often each tagger emerges as the most optimal choice for a specific metric. I.e. how often a tagger is on the skyline for a given size and accuracy metric.

With memory usage SVMTool dominates all taggers with lower accuracy, as none of them have a footprint that is smaller. When more memory is available, BiLSTM (Plank) or Flair stands out as the two best choices. Flair reaches highest token accuracy for seven of the eight languages, but requires more memory than the Plank model. Even though BERT-BPEmb has a performance almost at Flair's level, it only appears on the skyline for Chinese. This is because Flair not only has a better performance, but also a smaller footprint.

When it comes to model size and compressed model size, Brill and HMM are not dominated in the same fashion as for memory consumption. HMM is efficient for all languages, but neither reach strong peak accuracy. The HMM model from NLTK performs well in this metric because of its small size. SVMTool is efficient for all languages when it is compressed.

Looking at plots from Figure \ref{fig:skyline_rankings_token} and \ref{fig:skyline_rankings_sentence}, SVMTool, BiLSTM (Plank), and Flair are the most generally efficient taggers. They achieve optimal tradeoff in size vs. accuracy across most metrics for the tested taggers. SVMTool is the most versatile choice to use on embedded devices. It keeps a good balance of small size across metrics with decent accuracy. While the remaining classical taggers are more lightweight in certain size metrics, they do not manage the same consistent accuracy as SVMTool.

These results show that a larger and more complex model does not guarantee consistent performance. Flair has the highest consistent accuracy across languages, but is not the largest model for any size metrics. The same is true for SVMTool. It outperforms all other classical taggers except Stanford Tagger in terms of performance. Stanford Tagger is larger than SVMTool and only support a small set of languages.

%% file: sections/5_discussion.tex
\section{Discussion}
Among the deep learning taggers, BiLSTM (Plank) and Flair are the ones that are most often efficient, seen in Figures~\ref{fig:skyline_rankings_token} and~\ref{fig:skyline_rankings_sentence}. If one was to compress contemporary PoS taggers for use on constrained devices, these two taggers are obvious candidates. For example, using an aggressive compression pipeline with pruning, quantization, and weight sharing, as done by \newcite{Han2016DeepCoding}.

All deep learning taggers use some form of embeddings. For instance, most of Flair's size is due to the dimensions of the FastText embeddings. Compression of embeddings could help lower the memory and disk space required by the deep learning taggers. One could examine the approach by \newcite{Kim2020AdaptiveEmbeddings}, where they learn NLP task-specific features and compress embeddings accordingly. A different approach by \newcite{Liao2020EmbeddingQuantization} used quantization and dimensionality reduction to optimize the encoding of word embeddings, while preserving relevant information.

\subsection{Limitations}

\paragraph{Code Base Measurement}
As a fourth size metric, we considered measuring the size of the code base underlying each tagger. Estimating this is not trivial. The taggers from our experiments are written in various high-level languages and use various frameworks. A lot of the actual code logic is abstracted away in dependency chains. This obfuscates the actual size of the code required to run the taggers. 

Production code is often rewritten to be more compact, without general purpose frameworks, and in a low-level language. This makes measurement of high-level language code less interesting.

\paragraph{Visualisation}
In our experiments we trained ten PoS taggers on eight different languages. We measured two accuracy metrics and four size metrics. Creating plots for every combination of metrics would result in 48 individual plots. This would be confusing and not add value. Merely taking a tagger's average accuracy across all languages would obfuscate interesting information: Languages that have different characteristics are not directly comparable. Additionally, datasets for languages had varying sizes. We limited the featured skyline plots to a combination of token prediction accuracy and memory footprint. Token accuracy is the most common accuracy metric for PoS tagging. Memory usage is interesting because it captures the effective usage of the code base and model parameters at runtime, where memory is often a bottleneck on resource-constrained devices.

%% file: sections/6_conclusion.tex
\section{Conclusion}
This paper addressed how to measure the size of NLP models and how to select tools that optimise both predictive performance and size. We presented multiple size metrics, and a multi-lingual comparison of a broad range of part-of-speech taggers. We have measured their token and sentence accuracy, along with their size footprints. When combining these two groups of metrics, size and accuracy, we have shown how to determine which taggers are efficient, and which are not. 




The methods introduced give means for selecting optimal tool under either size or performance constraints (or both), as well as identifying which tools present no efficiency. Further, the results show that learning and encoding general knowledge, from a wide array of languages and datasets, is possible even with a small size footprint.

\newpage

%% file: sections/8_references.tex
\bibliographystyle{packages/acl_natbib}
\bibliography{mendeley,additions}

%% file: main.bbl
\begin{thebibliography}{35}
\expandafter\ifx\csname natexlab\endcsname\relax\def\natexlab#1{#1}\fi

\bibitem[{Akbik et~al.(2018)Akbik, Blythe, and
  Vollgraf}]{Akbik2018ContextualLabeling}
Alan Akbik, Duncan Blythe, and Roland Vollgraf. 2018.
\newblock {Contextual String Embeddings for Sequence Labeling}.
\newblock \emph{Proceedings of the 27th International Conference on
  Computational Linguistics}.

\bibitem[{Al-Rfou et~al.(2013)Al-Rfou, Perozzi, and
  Skiena}]{Al-Rfou2013Polyglot:NLP}
Rami Al-Rfou, Bryan Perozzi, and Steven Skiena. 2013.
\newblock {Polyglot: Distributed word representations for multilingual NLP}.
\newblock In \emph{CoNLL 2013 - 17th Conference on Computational Natural
  Language Learning, Proceedings}.

\bibitem[{Ardakani et~al.(2019)Ardakani, Ji, Smithson, Meyer, and
  Gross}]{Ardakani2019LearningWeights}
Arash Ardakani, Zhengyun Ji, Sean~C. Smithson, Brett~H. Meyer, and Warren~J.
  Gross. 2019.
\newblock {Learning recurrent binary/ternary weights}.
\newblock In \emph{7th International Conference on Learning Representations,
  ICLR 2019}.

\bibitem[{Bender et~al.(2021)Bender, Gebru, McMillan-Major, and
  Shmitchell}]{bender2021dangers}
Emily~M Bender, Timnit Gebru, Angelina McMillan-Major, and Shmargaret
  Shmitchell. 2021.
\newblock On the dangers of stochastic parrots: {C}an language models be too
  big? \raisebox{-5pt}{\includegraphics[scale=0.1]{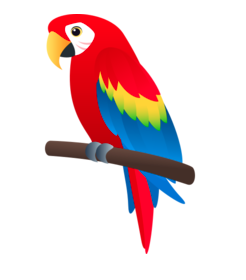}}.
\newblock \emph{Proceedings of FAccT}.

\bibitem[{Bird et~al.(2002)Bird, Bird, and Loper}]{Bird2002NLTKToolkit}
Steven Bird, Steven Bird, and Edward Loper. 2002.
\newblock {NLTK : The natural language toolkit NLTK : The Natural Language
  Toolkit}.
\newblock \emph{Proceedings of the ACL-02 Workshop on Effective tools and
  methodologies for teaching natural language processing and computational
  linguistics-Volume 1}.

\bibitem[{Bohnet et~al.(2018)Bohnet, McDonald, Simões, Andor, Pitler, and
  Maynez}]{Bohnet2018MorphosyntacticEncodings}
Bernd Bohnet, Ryan McDonald, Gonçalo Simões, Daniel Andor, Emily Pitler, and
  Joshua Maynez. 2018.
\newblock \href {https://doi.org/10.18653/v1/p18-1246} {{Morphosyntactic
  tagging with a meta-bilSTM model over context sensitive token encodings}}.
\newblock In \emph{ACL 2018 - 56th Annual Meeting of the Association for
  Computational Linguistics, Proceedings of the Conference (Long Papers)}.

\bibitem[{B{\"{o}}rzs{\"{o}}nyi et~al.(2001)B{\"{o}}rzs{\"{o}}nyi, Kossmann,
  and Stocker}]{Borzsonyi2001TheOperator}
S.~B{\"{o}}rzs{\"{o}}nyi, D.~Kossmann, and K.~Stocker. 2001.
\newblock \href {https://doi.org/10.1109/icde.2001.914855} {{The skyline
  operator}}.
\newblock In \emph{Proceedings - International Conference on Data Engineering}.

\bibitem[{Brants(2000)}]{Brants2000TnTTagger}
Thorsten Brants. 2000.
\newblock \href {https://arxiv.org/abs/cs/0003055} {{TnT - A Statistical
  Part-of-Speech Tagger}}.
\newblock \emph{CoRR}, cs.CL/0003.

\bibitem[{Brill(1992)}]{Brill1992ATagger}
Eric Brill. 1992.
\newblock \href {https://doi.org/10.3115/974499.974526} {{A simple rule-based
  part of speech tagger}}.
\newblock In \emph{ANLC '92: Proceedings of the third conference on Applied
  natural language processing}, page 152.

\bibitem[{Brown et~al.(2020)Brown, Mann, Ryder, Subbiah, Kaplan, Dhariwal,
  Neelakantan, Shyam, Sastry, Askell, Agarwal, Herbert-Voss, Krueger, Henighan,
  Child, Ramesh, Ziegler, Wu, Winter, Hesse, Chen, Sigler, Litwin, Gray, Chess,
  Clark, Berner, McCandlish, Radford, Sutskever, and
  Amodei}]{Brown2020LanguageLearners}
Tom~B. Brown, Benjamin Mann, Nick Ryder, Melanie Subbiah, Jared Kaplan,
  Prafulla Dhariwal, Arvind Neelakantan, Pranav Shyam, Girish Sastry, Amanda
  Askell, Sandhini Agarwal, Ariel Herbert-Voss, Gretchen Krueger, Tom Henighan,
  Rewon Child, Aditya Ramesh, Daniel~M. Ziegler, Jeffrey Wu, Clemens Winter,
  Christopher Hesse, Mark Chen, Eric Sigler, Mateusz Litwin, Scott Gray,
  Benjamin Chess, Jack Clark, Christopher Berner, Sam McCandlish, Alec Radford,
  Ilya Sutskever, and Dario Amodei. 2020.
\newblock {Language models are few-shot learners}.

\bibitem[{Buys and Botha(2016)}]{Buys2016Cross-lingualLanguages}
Jan Buys and Jan~A. Botha. 2016.
\newblock \href {https://doi.org/10.18653/v1/p16-1184} {{Cross-lingual
  morphological tagging for low-resource languages}}.
\newblock In \emph{54th Annual Meeting of the Association for Computational
  Linguistics, ACL 2016 - Long Papers}.

\bibitem[{Cardenas et~al.(2019)Cardenas, Lin, Ji, and
  May}]{Cardenas2019ALanguages}
Ronald Cardenas, Ying Lin, Heng Ji, and Jonathan May. 2019.
\newblock \href {https://doi.org/10.18653/v1/n19-1252} {{A grounded
  unsupervised universal part-of-speech tagger for low-resource languages}}.
\newblock In \emph{NAACL HLT 2019 - 2019 Conference of the North American
  Chapter of the Association for Computational Linguistics: Human Language
  Technologies - Proceedings of the Conference}.

\bibitem[{Chen et~al.(2016)Chen, Mou, Xu, Li, and
  Jin}]{Chen2016CompressingRepresentations}
Yunchuan Chen, Lili Mou, Yan Xu, Ge~Li, and Zhi Jin. 2016.
\newblock \href {https://doi.org/10.18653/v1/p16-1022} {{Compressing neural
  language models by sparse word representations}}.
\newblock In \emph{54th Annual Meeting of the Association for Computational
  Linguistics, ACL 2016 - Long Papers}.

\bibitem[{Devlin et~al.(2019)Devlin, Chang, Lee, and
  Toutanova}]{Devlin2019BERT:Understanding}
Jacob Devlin, Ming~Wei Chang, Kenton Lee, and Kristina Toutanova. 2019.
\newblock {BERT: Pre-training of deep bidirectional transformers for language
  understanding}.
\newblock In \emph{NAACL HLT 2019 - 2019 Conference of the North American
  Chapter of the Association for Computational Linguistics: Human Language
  Technologies - Proceedings of the Conference}.

\bibitem[{Ezen-Can(2020)}]{Ezen-Can2020ACorpus}
Aysu Ezen-Can. 2020.
\newblock {A comparison of LSTM and BERT for small corpus}.

\bibitem[{Fedus et~al.(2021)Fedus, Zoph, and Shazeer}]{fedus2021switch}
William Fedus, Barret Zoph, and Noam Shazeer. 2021.
\newblock Switch transformers: Scaling to trillion parameter models with simple
  and efficient sparsity.
\newblock \emph{arXiv preprint arXiv:2101.03961}.

\bibitem[{Gim{\'{e}}nez and M{\`{a}}rquez(2004)}]{Gimenez2004SVMTool:Machines}
Jesús Gim{\'{e}}nez and Lluís M{\`{a}}rquez. 2004.
\newblock {SVMTool: A general POS tagger generator based on support vector
  machines}.
\newblock In \emph{Proceedings of the 4th International Conference on Language
  Resources and Evaluation, LREC 2004}.

\bibitem[{Gupta et~al.(2015)Gupta, Agrawal, Gopalakrishnan, and
  Narayanan}]{Gupta2015DeepPrecision}
Suyog Gupta, Ankur Agrawal, Kailash Gopalakrishnan, and Pritish Narayanan.
  2015.
\newblock {Deep learning with limited numerical precision}.
\newblock In \emph{32nd International Conference on Machine Learning, ICML
  2015}.

\bibitem[{Han et~al.(2016)Han, Mao, and Dally}]{Han2016DeepCoding}
Song Han, Huizi Mao, and William~J. Dally. 2016.
\newblock {Deep compression: Compressing deep neural networks with pruning,
  trained quantization and Huffman coding}.
\newblock In \emph{4th International Conference on Learning Representations,
  ICLR 2016 - Conference Track Proceedings}.

\bibitem[{Heinzerling and Strube(2019)}]{Heinzerling2019BPEMB:Languages}
Benjamin Heinzerling and Michael Strube. 2019.
\newblock {BPEMB: Tokenization-free pre-trained subword embeddings in 275
  languages}.
\newblock In \emph{LREC 2018 - 11th International Conference on Language
  Resources and Evaluation}.

\bibitem[{Heinzerling and Strube(2020)}]{Heinzerling2020SequenceEvaluation}
Benjamin Heinzerling and Michael Strube. 2020.
\newblock \href {https://doi.org/10.18653/v1/p19-1027} {{Sequence tagging with
  contextual and non-contextual subword representations: A multilingual
  evaluation}}.
\newblock In \emph{ACL 2019 - 57th Annual Meeting of the Association for
  Computational Linguistics, Proceedings of the Conference}.

\bibitem[{Ignatov and Ignatov(2020)}]{Ignatov2020ControllingNetwork}
Dmitry Ignatov and Andrey Ignatov. 2020.
\newblock \href {https://doi.org/10.1016/j.patrec.2020.07.033} {{Controlling
  information capacity of binary neural network}}.
\newblock \emph{Pattern Recognition Letters}.

\bibitem[{Joulin et~al.(2017)Joulin, Grave, Bojanowski, and
  Mikolov}]{Joulin2017BagClassification}
Armand Joulin, Edouard Grave, Piotr Bojanowski, and Tomas Mikolov. 2017.
\newblock \href {https://doi.org/10.18653/v1/e17-2068} {{Bag of tricks for
  efficient text classification}}.
\newblock In \emph{15th Conference of the European Chapter of the Association
  for Computational Linguistics, EACL 2017 - Proceedings of Conference}.

\bibitem[{Kim et~al.(2020)Kim, Kim, and Lee}]{Kim2020AdaptiveEmbeddings}
Yeachan Kim, Kang-Min Kim, and SangKeun Lee. 2020.
\newblock \href {https://doi.org/10.18653/v1/2020.acl-main.364} {{Adaptive
  Compression of Word Embeddings}}.
\newblock In \emph{ACL - 58th Annual Meeting of the Association for
  Computational Linguistics, Proceedings of the Conference (Long Papers)}.

\bibitem[{Liao et~al.(2020)Liao, Chen, Wang, Qiu, and
  Yuan}]{Liao2020EmbeddingQuantization}
Siyu Liao, Jie Chen, Yanzhi Wang, Qinru Qiu, and Bo~Yuan. 2020.
\newblock \href {https://doi.org/10.1609/aaai.v34i05.6350} {{Embedding
  compression with isotropic iterative quantization}}.

\bibitem[{Mahoney(2003)}]{barf}
Matt Mahoney. 2003.
\newblock Better archiver with recursive functionality ({BARF}).
\newblock http://mattmahoney.net/dc/barf.html.

\bibitem[{Marcus et~al.(1993)Marcus, Santorini, and
  Marcinkiewicz}]{Marcus1993BuildingTreebank}
M.~Marcus, B~Santorini, and M.~Marcinkiewicz. 1993.
\newblock {Building a Large Annotated Corpus of English: The Penn Treebank}.
\newblock \emph{Computational linguistics - Association for Computational
  Linguistics (Print)}.

\bibitem[{Melas-Kyriazi et~al.(2020)Melas-Kyriazi, Han, and
  Liang}]{Melas-Kyriazi2020Generation-distillationSettings}
Luke Melas-Kyriazi, George Han, and Celine Liang. 2020.
\newblock \href {https://doi.org/10.18653/v1/d19-6114}
  {{Generation-distillation for efficient natural language understanding in
  low-data settings}}.

\bibitem[{Nivre et~al.(2020)Nivre, de~Marneffe, Ginter, Haji, Manning, Pyysalo,
  Schuster, and Zeman}]{Nivre2020UniversalCollection}
Joakim Nivre, Marie~Catherine de~Marneffe, Filip Ginter, Jan Haji,
  Christopher~D. Manning, Sampo Pyysalo, Sebastian Schuster, and Francis
  Tyers~Daniel Zeman. 2020.
\newblock {Universal dependencies v2: An evergrowing multilingual treebank
  collection}.
\newblock In \emph{LREC 2020 - 12th International Conference on Language
  Resources and Evaluation, Conference Proceedings}.

\bibitem[{Plank et~al.(2016)Plank, S{\o}gaard, and
  Goldberg}]{Plank2016MultilingualLoss}
Barbara Plank, Anders S{\o}gaard, and Yoav Goldberg. 2016.
\newblock \href {https://doi.org/10.18653/v1/p16-2067} {{Multilingual
  part-of-speech tagging with bidirectional long short-term memory models and
  auxiliary loss}}.
\newblock In \emph{54th Annual Meeting of the Association for Computational
  Linguistics, ACL 2016 - Short Papers}.

\bibitem[{Szymanski and Gorman(2020)}]{Szymanski2020IsProcessing}
Piotr Szymanski and Kyle Gorman. 2020.
\newblock \href {https://doi.org/10.18653/v1/2020.emnlp-main.172} {{Is the best
  better? Bayesian statistical model comparison for natural language
  processing}}.

\bibitem[{Tissier et~al.(2019)Tissier, Gravier, and
  Habrard}]{Tissier2019Near-losslessEmbeddings}
Julien Tissier, Christophe Gravier, and Amaury Habrard. 2019.
\newblock \href {https://doi.org/10.1609/aaai.v33i01.33017104} {{Near-lossless
  binarization of word embeddings}}.
\newblock In \emph{33rd AAAI Conference on Artificial Intelligence, AAAI 2019,
  31st Innovative Applications of Artificial Intelligence Conference, IAAI 2019
  and the 9th AAAI Symposium on Educational Advances in Artificial
  Intelligence, EAAI 2019}.

\bibitem[{Toutanova et~al.(2003)Toutanova, Klein, Manning, and
  Singer}]{Toutanova2003Feature-richNetwork}
Kristina Toutanova, Dan Klein, Christopher~D. Manning, and Yoram Singer. 2003.
\newblock \href {https://doi.org/10.3115/1073445.1073478} {{Feature-rich
  part-of-speech tagging with a cyclic dependency network}}.
\newblock In \emph{In Proceedings of HLT-NAACL 2003}, pages 173--180.

\bibitem[{Wang et~al.(2018)Wang, Choi, Brand, Chen, and
  Gopalakrishnan}]{Wang2018TrainingNumbers}
Naigang Wang, Jungwook Choi, Daniel Brand, Chia~Yu Chen, and Kailash
  Gopalakrishnan. 2018.
\newblock {Training deep neural networks with 8-bit floating point numbers}.
\newblock In \emph{Advances in Neural Information Processing Systems}.

\bibitem[{Yasunaga et~al.(2018)Yasunaga, Kasai, and
  Radev}]{Yasunaga2018RobustTraining}
Michihiro Yasunaga, Jungo Kasai, and Dragomir Radev. 2018.
\newblock \href {https://doi.org/10.18653/v1/n18-1089} {{Robust multilingual
  part-of-speech tagging via adversarial training}}.
\newblock In \emph{NAACL HLT 2018 - 2018 Conference of the North American
  Chapter of the Association for Computational Linguistics: Human Language
  Technologies - Proceedings of the Conference}.

\end{thebibliography}
